\documentclass[conference]{IEEEtran}
\usepackage{times}
\usepackage[nolist]{acronym} 
\usepackage{graphicx} 
\usepackage[numbers]{natbib}
\usepackage{multicol}
\usepackage[bookmarks=true]{hyperref}
\usepackage{pgfplots}
\usepackage{float}
\usepackage{dblfloatfix}
\usepackage{xcolor}
\usepackage{subfig}
\graphicspath{{figs/}}

\definecolor{findOptimalPartition}{HTML}{D7191C}
\definecolor{storeClusterComponent}{HTML}{FDAE61}
\definecolor{dbscan}{HTML}{ABDDA4}
\definecolor{constructCluster}{HTML}{2B83BA}
\usetikzlibrary{patterns}
\pgfplotsset{compat=1.16} 
\newcommand{\figref}[1]{\hyperref[#1]{Fig.~\ref*{#1}}}
\newcommand{\tabref}[1]{\hyperref[#1]{Table~\ref*{#1}}}
\newcommand{\secref}[1]{\hyperref[#1]{Section~\ref*{#1}}}
\newcommand{\algoref}[1]{\hyperref[#1]{Algorithm~\ref*{#1}}}

\def\ie{\textit{i.e.,}}

\newacro{mss}[MSS]{Mass-spring system}
\newacro{pbd}[PBD]{Position-based dynamics}
\newacro{fem}[FEM]{Finite element method}
\newacro{dnn}[DNN]{Deep Neural Network}

\begin{document}

\title{Towards synthesizing grasps for 3D deformable objects with physics-based simulation}

\author{Author Names Omitted for Anonymous Review. Paper-ID 4}



%
\author{\authorblockN{Tran Nguyen Le,
Jens Lundell,
Fares J.Abu-Dakka, and
Ville Kyrki}
\authorblockA{Intelligent Robotics Group\\ Department of Electrical Engineering and Automation, School of Electrical Engineering, Aalto University, Finland}}

\maketitle


\IEEEpeerreviewmaketitle
\section{Introduction}
Despite recent remarkable successes in robotic grasping, most works on grasp synthesis assume either implicitly or explicitly rigid objects \cite{mahler2017dex,morrison2018closing,jens19}. Rigid objects simplify grasp planning to the choice of contact points along the object surface, but the assumption does not hold for many real objects. However, planning grasps on non-rigid objects is difficult because objects deform under interaction forces meaning that the 3-D contact locations also depend on the forces exerted on the object. Most existing works on planning grasps on deformable objects aim to minimize the deformation \cite{xu_minimalwork,pan_minimizedeform,delgado_minimize}, while some works \cite{lin_feel3d} actually take advantage of the deformation. Nevertheless, how object stiffness affects grasping and how to use the object deformation to generate grasps remains an open question. In this context, it is important to note that grasps generated by methods assuming rigid objects do not necessarily translate well to deformable objects and vice versa. Therefore, there is a need to study how to generate grasps that harness the target objects' stiffness. 

Of late, deep learning is the major driving force behind the progress in rigid object grasping. Many of these techniques share a similar pipeline, where a deep neural network is trained on a real or synthetic dataset to generate and evaluate grasp candidates given an input image. Rigid body simulators such as GraspIt! \cite{graspit} and OpenGRASP \cite{opengrasp} have been used to generate thousands of grasp candidates to serve as training data for those methods. The large amount of training data helped those methods to achieve remarkable successes in terms of grasp success rate on rigid objects. Recently, to study how to manipulate cloth and rope-type objects, simulators such as PyBullet \cite{pybullet} and MuJoCo \cite{mujoco} have been used \cite{wu_mujoco,yan_mujoco,jan_pybullet}. However, the use of these simulators for 3D solid deformable objects is still limited.

To address the aforementioned open issues, we envision an approach that generates grasps on a wider range of objects with varying stiffness by incorporating stiffness as an additional input to a state-of-the-art deep grasp planning pipeline (\figref{fig:pipeline}). Our system generates grasp candidates and grasp qualities for every pixel given an input depth image and stiffness image. When combined with depth information, the model outputs can be reprojected into 3D space, allowing a robot to execute a generated grasp in the real world.

The approach is evaluated in simulation and shows an improvement in terms of grasp success rate for a wide range of objects with various shapes and varying stiffness. The approach is able to generate different grasping strategies for different stiffness values such as pinching for soft objects and caging for hard objects even though no pinch grasps were included in the training data.
\vspace{-0.5 em}
\section{Grasp generation using physics-based simulation}
\vspace{-0.5em}
\subsection{Simulation platform choice}
Simulating dynamics of deformable objects relies heavily on their geometric representations. 
\citet{yin_survey} presents three primary deformable object modelling approaches, \ac{mss}, \ac{pbd}, and \ac{fem}, and their limitations. 
In this work, we use \ac{fem} because it is often used to model 3D objects such as food or tissues and, compared to other modeling approaches, offers a more physically accurate representation of a deformable object in a continuous domain.  

Most robotic simulators do not support \ac{fem} except NVIDIA's recent version of the Isaac Gym simulator \cite{isaacgym}, which supports soft body simulation through the NVIDIA Flex backend. 
Similar to SOFA \cite{sofa}, Isaac Gym includes co-rotational linear model for precision in modeling and simulating the object deformation under interaction. Furthermore, the Isaac simulator also provides the capability to integrate robot-related functions, making it easier to build robotic applications. 
NVIDIA also provides a grasping framework \cite{grasp_framework} to automatically perform and evaluate grasp tests on an arbitrary target object. We use this framework in our work to generate training data and test grasps.
\vspace{-0.5em}
\subsection{Grasp generation network}
To take object stiffness into account for generating grasps, we propose to use the \ac{dnn} (\figref{fig:pipeline}). The network is inspired by \cite{morrison2018closing} but modified to take a stiffness image as an additional input channel. Each pixel in the stiffness image represents object stiffness. The proposed network is trained with supervised learning on a synthetic dataset. We generated our own dataset containing labeled grasps on soft and rigid objects using Isaac Gym as no such dataset existed from before.
\vspace{-0.5em}
\subsection{Training data generation}
\textbf{Depth and stiffness input} We captured depth images of target objects with a virtual camera set to view the scene from top-down. To model variable object stiffness, four values of Young's modulus from $2\cdot 10^4$ to $2\cdot 10^9$ were used. The Young's modulus is normalized to [0,1] range and the corresponding stiffness value is assigned to every pixel in the stiffness image that the object occupies.

\textbf{Grasp candidates} Grasps are sampled with an antipodal grasp sampler to obtain approximately 200 grasp candidates for each target object. All grasp candidates that collide with the mesh are filtered out, a process that keeps about 25-40 grasps per object. The grasps are executed and evaluated using a Franka Emika Panda model in Isaac Gym. Positive grasps are then represented as rectangles in 2D image plane as shown in \figref{fig:grasprep}.  

\textbf{Quality metrics} None of the standard grasp quality metrics are applicable for both rigid and deformable objects. As a quality metric we use a shake task which measures how easily an object is displaced in hand under various accelerations. The metric is provided by the Isaac Gym framework. A higher metric indicates that a grasp is better because it withstand higher accelerations. 

\textbf{Training dataset} As a training dataset, we use a total of 30 objects on which we generate and label grasp candidates. The objects include 13 primitive objects provided in Isaac Gym, 5 objects from YCB dataset, and 12 objects with adversarial geometry from the EGAD! dataset~\cite{egad}. With the varying stiffness, the training set contains a total of 120 objects. To counteract the small size of the training set, we further augment the dataset with random crops, zooms, and rotations to create a set of 5400 depth and stiffness images and 27000 labeled grasps map images.  
\vspace{-0.5 em}
\section{Results}
We evaluated the quality of the proposed grasp generation in simulation on objects with varying stiffness. We tested the approach on 7 common objects shown in \figref{fig:testobj}. We evaluated the top-5 generated grasps using the shake test on each object for each of the four  stiffnesses, resulting in 20 grasps per object. To demonstrate the importance of stiffness input, we compared the generated grasps against grasps generated with a similar approach without stiffness information. 
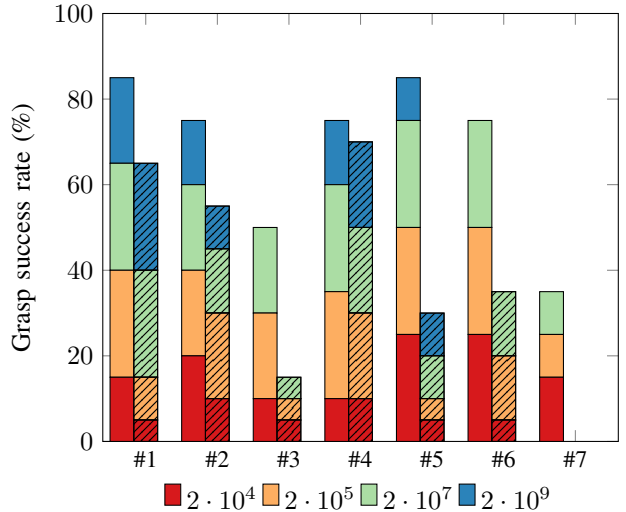
\begin{figure}
\centering
\begin{tikzpicture}[
/pgfplots/every axis/.style={ 
    ybar stacked,
    legend style={ legend columns=4,
    at={(xticklabel cs:0.5)},
    anchor=north,
    draw=none}, 
    symbolic x coords={
      \#1, \#2,\#3, \#4, \#5, \#6, \#7},
    bar width=9pt,
    ylabel={Grasp success rate (\%)},
    ymin=0,
    ymax=100,
    xtick=data,
    x tick label style={font=\small, 
    },
    point meta=rawy,
  },
]

\begin{axis}[bar shift=-9pt] 
\addplot [black,fill=findOptimalPartition] coordinates {(\#1,15) (\#2,20) (\#3,10) (\#4, 10) (\#5, 25) (\#6,25) (\#7, 15)};
\addlegendentry[]{$2\cdot 10^4$};
\addplot [black,fill=storeClusterComponent] coordinates {(\#1,25) (\#2,20) (\#3,20) (\#4, 25) (\#5, 25) (\#6,25) (\#7, 10)};
\addlegendentry[]{$2\cdot 10^5$};
\addplot [black,fill=dbscan] coordinates {(\#1,25) (\#2,20) (\#3,20) (\#4, 25) (\#5, 25) (\#6,25) (\#7, 10)};
\addlegendentry[]{$2\cdot 10^7$};
\addplot [black,fill=constructCluster] coordinates {(\#1,20) (\#2,15) (\#3,0) (\#4, 15) (\#5, 10) (\#6,0) (\#7, 0)};
\addlegendentry[]{$2\cdot 10^9$};
\end{axis}

\begin{axis}[bar shift=0pt,hide axis]
\addplot+[draw = black,fill=findOptimalPartition, postaction={
        pattern=north east lines
    }]  coordinates {(\#1,5) (\#2,10) (\#3,5) (\#4, 10) (\#5, 5) (\#6,5) (\#7, 0)};

\addplot+[draw = black, fill=storeClusterComponent, postaction={
        pattern=north east lines
    }]  coordinates {(\#1,10) (\#2,20) (\#3,5) (\#4, 20) (\#5, 5) (\#6,15) (\#7, 0)};
\addplot+[draw = black,fill=dbscan, postaction={
        pattern=north east lines
    }]  coordinates {(\#1,25) (\#2,15) (\#3,5) (\#4, 20) (\#5, 10) (\#6,15) (\#7, 0)};
\addplot+[draw = black,fill=constructCluster, postaction={
        pattern=north east lines
    }]  coordinates {(\#1,25) (\#2,10) (\#3,0) (\#4, 20) (\#5, 10) (\#6,0) (\#7, 0)};
\end{axis}
\end{tikzpicture}
\caption{Grasp success rate of 7 common objects with four values of Young's modulus. Plain columns present the result of the model that takes stiffness input into account and striped columns present that of the model that use only depth image as input.}
\label{fig:result_7}
\vspace{-1.5em}
\end{figure}

Over all stiffnesses, the average grasp success rate is 30\% higher with the proposed approach that takes stiffness input into account compared to the baseline where stiffness input is ignored. This result stems from the fact that the performance of the baseline approach deteriorates significantly when changing from high value to low value of the Young's modulus. For instance, the relative performance drop for the baseline approach from $2\cdot 10^7$ to $2\cdot 10^5$ is 10\% and to $2\cdot 10^4$ the drop is 30\%. This is much higher compared to the 0\% and 12\% drop using our approach. As the baseline approach does not consider stiffness input, that method generates the same grasps for a target object regardless its stiffness. These generated grasps may successfully grasp the objects, however, during the shake task, the objects usually slip away from the gripper due to their deformation. Therefore, by taking the stiffness input into account, the proposed network is able to learn to avoid areas with high probability of slippage, resulting in higher grasp success rate.


Another interesting finding is that our approach can generate different grasp types such as pinch or cage grasps depending on the stiffness, even though there were no pinch grasps in the training dataset. This behavior is shown on object 5 in \figref{fig:demo}. Specifically, the sponge with a low Young's modulus admits pinching behavior where the grasp press on the object and pinch, while the hard sponge only admits caging grasps. One possible reason behind this behavior is that the proposed network learned that the grasp quality is high almost everywhere on soft objects thanks to their deformation. Similar behaviour was also reported in \cite{sergey_handeye} on a few objects. It is worth pointing out that our approach produces the same behavior as in \cite{sergey_handeye} but on a completely synthetic dataset containing order of magnitudes less data. Furthermore, our proposed approach provides more meaningful insights in terms of the relationship between object deformation and grasps. 
\vspace{-0.5 em}
\section{Conclusion}
Grasping deformable objects is not well researched due to complexity in the modelling and simulating the dynamic behavior of such objects. However, with the rapid development of physics-based simulators that support soft bodies, the research gap between rigid and deformable objects is getting smaller. To leverage the capability of such simulators and to challenge the assumption that has guided robotic grasping research so far, \ie, object rigidity, we proposed a deep-learning based approach that generates stiffness dependent grasps. Our network is trained on purely synthetic data generated from a physics-based simulator. The same simulator is also used to evaluate the trained network. The results show improvement in terms of grasp ranking and grasp success rate. Furthermore, our network can adapt the grasps based on the stiffness. We are currently validating the proposed approach on a larger test dataset in simulation and on a physical robot.   



\bibliographystyle{plainnat}
\bibliography{references}

\begin{thebibliography}{20}
\providecommand{\natexlab}[1]{#1}
\providecommand{\url}[1]{\texttt{#1}}
\expandafter\ifx\csname urlstyle\endcsname\relax
  \providecommand{\doi}[1]{doi: #1}\else
  \providecommand{\doi}{doi: \begingroup \urlstyle{rm}\Url}\fi

\bibitem[Coumans and Bai(2016--2021)]{pybullet}
Erwin Coumans and Yunfei Bai.
\newblock Pybullet, a python module for physics simulation for games, robotics
  and machine learning.
\newblock \url{http://pybullet.org}, 2016--2021.

\bibitem[Delgado et~al.(2015)Delgado, Jara, Mira, and Torres]{delgado_minimize}
A.~Delgado, C.~A. Jara, D.~Mira, and F.~Torres.
\newblock \href{https://ieeexplore.ieee.org/document/7347794}{Tactile-based
  grasping strategy for deformable objects' manipulation and deformability
  estimation}.
\newblock In \emph{2015 12th International Conference on Informatics in
  Control, Automation and Robotics (ICINCO)}, volume~02, pages 369--374, 2015.

\bibitem[Faure et~al.(2012)Faure, Duriez, Delingette, Allard, Gilles,
  Marchesseau, Talbot, Courtecuisse, Bousquet, Peterlik, and Cotin]{sofa}
Fran{\c c}ois Faure, Christian Duriez, Herv{\'e} Delingette, J{\'e}r{\'e}mie
  Allard, Benjamin Gilles, St{\'e}phanie Marchesseau, Hugo Talbot, Hadrien
  Courtecuisse, Guillaume Bousquet, Igor Peterlik, and St{\'e}phane Cotin.
\newblock \href{https://link.springer.com/chapter/10.1007/8415_2012_125}{SOFA:
  A Multi-Model Framework for Interactive Physical Simulation}.
\newblock In Yohan Payan, editor, \emph{{Soft Tissue Biomechanical Modeling for
  Computer Assisted Surgery}}, volume~11 of \emph{Studies in Mechanobiology,
  Tissue Engineering and Biomaterials}, pages 283--321. {Springer}, June 2012.

\bibitem[Le{\'o}n et~al.(2010)Le{\'o}n, Ulbrich, Diankov, Puche, Przybylski,
  Morales, Asfour, Moisio, Bohg, Kuffner, and Dillmann]{opengrasp}
Beatriz Le{\'o}n, Stefan Ulbrich, Rosen Diankov, Gustavo Puche, Markus
  Przybylski, Antonio Morales, Tamim Asfour, Sami Moisio, Jeannette Bohg, James
  Kuffner, and R{\"u}diger Dillmann.
\newblock
  \href{https://link.springer.com/chapter/10.1007/978-3-642-17319-6_13}{OpenGRASP:
  A Toolkit for Robot Grasping Simulation}.
\newblock In Noriaki Ando, Stephen Balakirsky, Thomas Hemker, Monica Reggiani,
  and Oskar von Stryk, editors, \emph{Simulation, Modeling, and Programming for
  Autonomous Robots}, pages 109--120, Berlin, Heidelberg, 2010. Springer Berlin
  Heidelberg.
\newblock ISBN 978-3-642-17319-6.

\bibitem[Levine et~al.(2018)Levine, Pastor, Krizhevsky, Ibarz, and
  Quillen]{sergey_handeye}
Sergey Levine, Peter Pastor, Alex Krizhevsky, Julian Ibarz, and Deirdre
  Quillen.
\newblock
  \href{https://journals.sagepub.com/doi/full/10.1177/0278364917710318}{Learning
  hand-eye coordination for robotic grasping with deep learning and large-scale
  data collection}.
\newblock \emph{The International Journal of Robotics Research}, 37\penalty0
  (4-5):\penalty0 421--436, 2018.

\bibitem[Lin et~al.(2015)Lin, Guo, Wang, and Jia]{lin_feel3d}
Huan Lin, Feng Guo, Feifei Wang, and Yan-Bin Jia.
\newblock
  \href{https://journals.sagepub.com/doi/abs/10.1177/0278364914564232?journalCode=ijra}{Picking
  up a soft 3D object by “feeling” the grip}.
\newblock \emph{The International Journal of Robotics Research}, 34\penalty0
  (11):\penalty0 1361--1384, 2015.

\bibitem[Lundell et~al.(2019)Lundell, Verdoja, and Kyrki]{jens19}
Jens Lundell, Francesco Verdoja, and Ville Kyrki.
\newblock \href{https://ieeexplore.ieee.org/document/8967816}{Robust Grasp
  Planning Over Uncertain Shape Completions}.
\newblock In \emph{2019 IEEE/RSJ International Conference on Intelligent Robots
  and Systems (IROS)}, pages 1526--1532, 2019.
\newblock \doi{10.1109/IROS40897.2019.8967816}.

\bibitem[Mahler et~al.(2017)Mahler, Liang, Niyaz, Laskey, Doan, Liu, Ojea, and
  Goldberg]{mahler2017dex}
Jeffrey Mahler, Jacky Liang, Sherdil Niyaz, Michael Laskey, Richard Doan, Xinyu
  Liu, Juan~Aparicio Ojea, and Ken Goldberg.
\newblock \href{https://berkeleyautomation.github.io/dex-net/#dexnet_2}{Dex-Net
  2.0: Deep Learning to Plan Robust Grasps with Synthetic Point Clouds and
  Analytic Grasp Metrics}.
\newblock 2017.

\bibitem[Matas et~al.(2018)Matas, James, and Davison]{jan_pybullet}
Jan Matas, Stephen James, and Andrew~J. Davison.
\newblock Sim-to-real reinforcement learning for deformable object
  manipulation.
\newblock In \emph{2nd Annual Conference on Robot Learning, CoRL 2018,
  Z{\"{u}}rich, Switzerland, 29-31 October 2018, Proceedings}, volume~87 of
  \emph{Proceedings of Machine Learning Research}, pages 734--743. {PMLR},
  2018.
\newblock URL \url{http://proceedings.mlr.press/v87/matas18a.html}.

\bibitem[Miller and Allen(2004)]{graspit}
A.T. Miller and P.K. Allen.
\newblock \href{https://ieeexplore.ieee.org/document/1371616}{Graspit! A
  versatile simulator for robotic grasping}.
\newblock \emph{IEEE Robotics Automation Magazine}, 11\penalty0 (4):\penalty0
  110--122, 2004.

\bibitem[Morrison et~al.(2018)Morrison, Corke, and
  Leitner]{morrison2018closing}
Douglas Morrison, Peter Corke, and J\"urgen Leitner.
\newblock \href{http://roboticsproceedings.org/rss14/p21.pdf}{Closing the Loop
  for Robotic Grasping: A Real-time, Generative Grasp Synthesis Approach}.
\newblock In \emph{Proc.\ of Robotics: Science and Systems (RSS)}, 2018.

\bibitem[Morrison et~al.(2020)Morrison, Corke, and Leitner]{egad}
Douglas Morrison, Peter Corke, and Jürgen Leitner.
\newblock \href{https://ieeexplore.ieee.org/document/9085936}{EGAD! An Evolved
  Grasping Analysis Dataset for Diversity and Reproducibility in Robotic
  Manipulation}.
\newblock \emph{IEEE Robotics and Automation Letters}, 5\penalty0 (3):\penalty0
  4368--4375, 2020.

\bibitem[NVIDIA(2020{\natexlab{a}})]{grasp_framework}
NVIDIA.
\newblock Deformable object grasping framework.
\newblock \url{https://github.com/NVlabs/deformable_object_grasping},
  2020{\natexlab{a}}.

\bibitem[NVIDIA(2020{\natexlab{b}})]{isaacgym}
NVIDIA.
\newblock Isaac gym.
\newblock \url{https://developer.nvidia.com/isaac-gym}, 2020{\natexlab{b}}.

\bibitem[Pan et~al.(2020)Pan, Gao, and Manocha]{pan_minimizedeform}
Zherong Pan, Xifeng Gao, and Dinesh Manocha.
\newblock \href{https://ieeexplore.ieee.org/document/9196938}{Grasping Fragile
  Objects Using A Stress-Minimization Metric}.
\newblock In \emph{2020 IEEE International Conference on Robotics and
  Automation (ICRA)}, pages 517--523, 2020.

\bibitem[Todorov et~al.(2012)Todorov, Erez, and Tassa]{mujoco}
Emanuel Todorov, Tom Erez, and Yuval Tassa.
\newblock \href{https://ieeexplore.ieee.org/document/6386109}{MuJoCo: A physics
  engine for model-based control}.
\newblock In \emph{2012 IEEE/RSJ International Conference on Intelligent Robots
  and Systems}, pages 5026--5033, 2012.

\bibitem[Wu et~al.(2020)Wu, Yan, Kurutach, Pinto, and Abbeel]{wu_mujoco}
Yilin Wu, Wilson Yan, Thanard Kurutach, Lerrel Pinto, and Pieter Abbeel.
\newblock {Learning to Manipulate Deformable Objects without Demonstrations}.
\newblock In \emph{Proceedings of Robotics: Science and Systems}, Corvalis,
  Oregon, USA, July 2020.
\newblock \doi{10.15607/RSS.2020.XVI.065}.

\bibitem[Xu et~al.(2020)Xu, Danielczuk, Ichnowski, Mahler, Steinbach, and
  Goldberg]{xu_minimalwork}
Jingyi Xu, Michael Danielczuk, Jeffrey Ichnowski, Jeffrey Mahler, Eckehard
  Steinbach, and Ken Goldberg.
\newblock \href{https://ieeexplore.ieee.org/abstract/document/9197062}{Minimal
  Work: A Grasp Quality Metric for Deformable Hollow Objects}.
\newblock In \emph{2020 IEEE International Conference on Robotics and
  Automation (ICRA)}, pages 1546--1552, 2020.

\bibitem[Yan et~al.(2020)Yan, Vangipuram, Abbeel, and Pinto]{yan_mujoco}
Wilson Yan, Ashwin Vangipuram, Pieter Abbeel, and Lerrel Pinto.
\newblock Learning predictive representations for deformable objects using
  contrastive estimation.
\newblock \emph{CoRR}, abs/2003.05436, 2020.
\newblock URL \url{https://arxiv.org/abs/2003.05436}.

\bibitem[Yin et~al.(2021)Yin, Varava, and Kragic]{yin_survey}
Hang Yin, Anastasia Varava, and Danica Kragic.
\newblock
  \href{https://robotics.sciencemag.org/content/6/54/eabd8803}{Modeling,
  learning, perception, and control methods for deformable object
  manipulation}.
\newblock \emph{Science Robotics}, 6\penalty0 (54), 2021.

\end{thebibliography}

\appendix
\counterwithin{figure}{section}

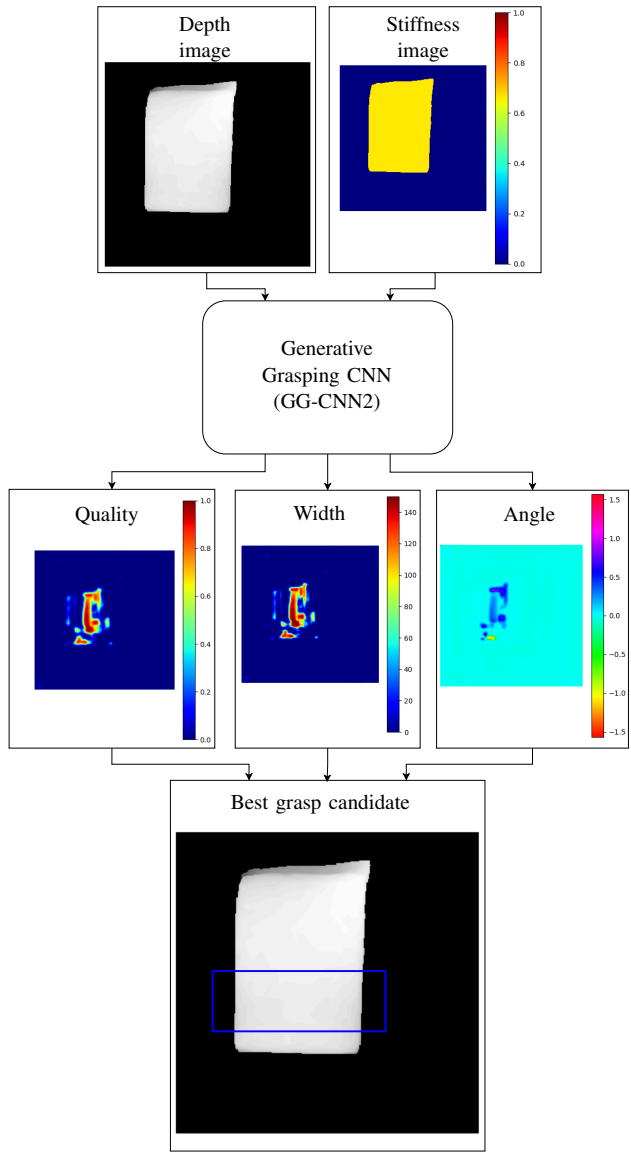
\begin{figure}[ht]
    \centering
	\def\svgwidth{\linewidth}
     {\fontsize{8}{8}
\begingroup%
  \makeatletter%
  \providecommand\color[2][]{%
    \errmessage{(Inkscape) Color is used for the text in Inkscape, but the package 'color.sty' is not loaded}%
    \renewcommand\color[2][]{}%
  }%
  \providecommand\transparent[1]{%
    \errmessage{(Inkscape) Transparency is used (non-zero) for the text in Inkscape, but the package 'transparent.sty' is not loaded}%
    \renewcommand\transparent[1]{}%
  }%
  \providecommand\rotatebox[2]{#2}%
  \newcommand*\fsize{\dimexpr\f@size pt\relax}%
  \newcommand*\lineheight[1]{\fontsize{\fsize}{#1\fsize}\selectfont}%
  \ifx\svgwidth\undefined%
    \setlength{\unitlength}{1239bp}%
    \ifx\svgscale\undefined%
      \relax%
    \else%
      \setlength{\unitlength}{\unitlength * \real{\svgscale}}%
    \fi%
  \else%
    \setlength{\unitlength}{\svgwidth}%
  \fi%
  \global\let\svgwidth\undefined%
  \global\let\svgscale\undefined%
  \makeatother%
  \begin{picture}(1,1.72094431)%
    \lineheight{1}%
    \setlength\tabcolsep{0pt}%
    \put(0,0){\includegraphics[width=\unitlength,page=1]{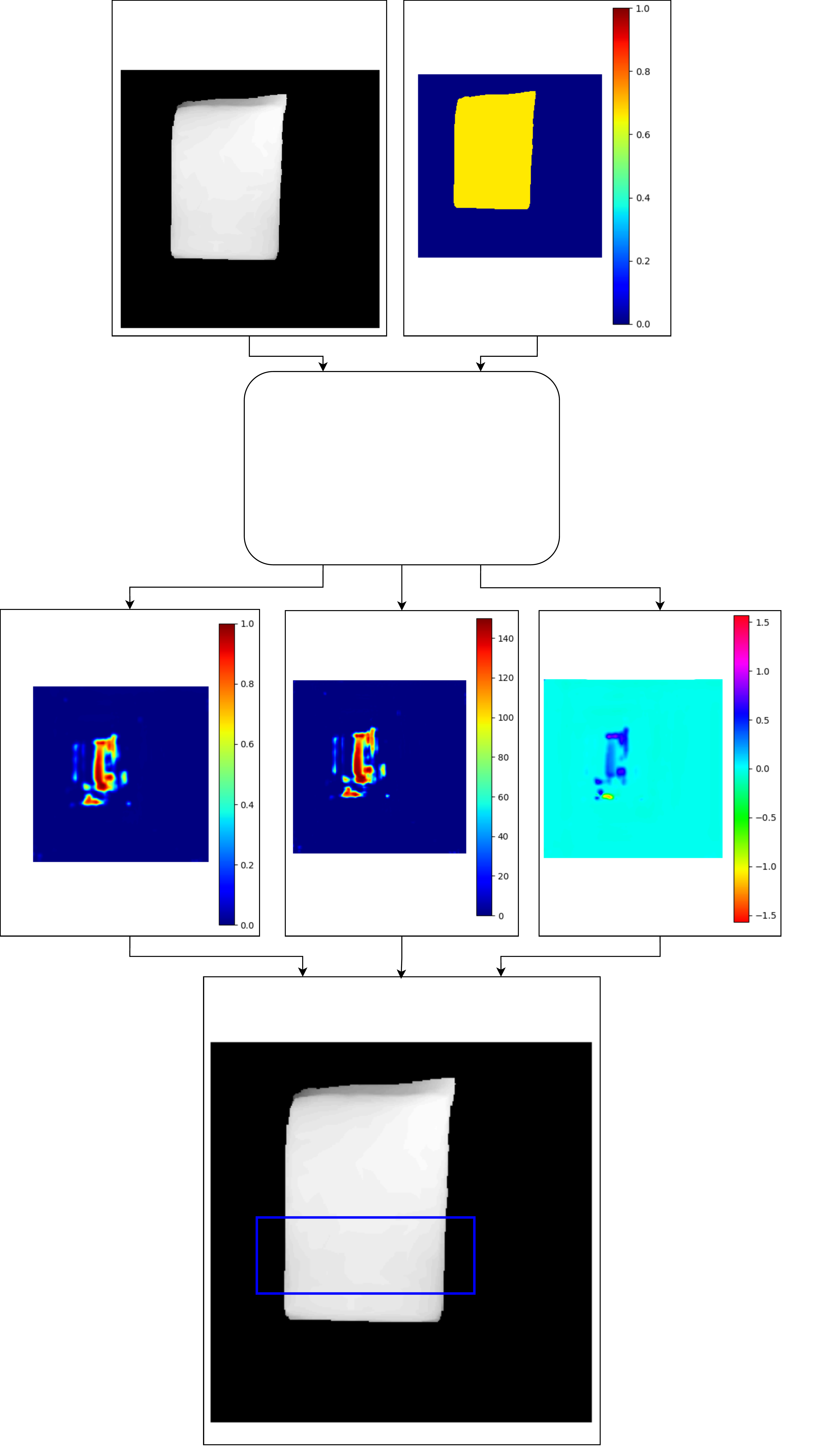}}%
    \put(0.29433035,1.68458984){\makebox(0,0)[t]{\lineheight{1.25}\smash{\begin{tabular}[t]{c}Depth \\image\end{tabular}}}}%
    \put(0.62182521,1.68398933){\makebox(0,0)[t]{\lineheight{1.25}\smash{\begin{tabular}[t]{c}Stiffness \\image\end{tabular}}}}%
    \put(0.47756219,1.19690307){\makebox(0,0)[t]{\lineheight{1.25}\smash{\begin{tabular}[t]{c}Generative\\Grasping CNN\\(GG-CNN2)\end{tabular}}}}%
    \put(0.14700804,0.9486384){\makebox(0,0)[t]{\lineheight{1.25}\smash{\begin{tabular}[t]{c}Quality\end{tabular}}}}%
    \put(0.46617788,0.94975344){\makebox(0,0)[t]{\lineheight{1.25}\smash{\begin{tabular}[t]{c}Width\end{tabular}}}}%
    \put(0.78119842,0.94862613){\makebox(0,0)[t]{\lineheight{1.25}\smash{\begin{tabular}[t]{c}Angle\end{tabular}}}}%
    \put(0.4674588,0.51607803){\makebox(0,0)[t]{\lineheight{1.25}\smash{\begin{tabular}[t]{c}Best grasp candidate\end{tabular}}}}%
  \end{picture}%
\endgroup%
}      
    \caption{The proposed pipeline where stiffness information is incorporated.}
    \label{fig:pipeline}
\end{figure}
\begin{figure}
    \centering
	\def\svgwidth{0.5\linewidth}
     {\color{white} \fontsize{8}{8}
\begingroup%
  \makeatletter%
  \providecommand\color[2][]{%
    \errmessage{(Inkscape) Color is used for the text in Inkscape, but the package 'color.sty' is not loaded}%
    \renewcommand\color[2][]{}%
  }%
  \providecommand\transparent[1]{%
    \errmessage{(Inkscape) Transparency is used (non-zero) for the text in Inkscape, but the package 'transparent.sty' is not loaded}%
    \renewcommand\transparent[1]{}%
  }%
  \providecommand\rotatebox[2]{#2}%
  \newcommand*\fsize{\dimexpr\f@size pt\relax}%
  \newcommand*\lineheight[1]{\fontsize{\fsize}{#1\fsize}\selectfont}%
  \ifx\svgwidth\undefined%
    \setlength{\unitlength}{459bp}%
    \ifx\svgscale\undefined%
      \relax%
    \else%
      \setlength{\unitlength}{\unitlength * \real{\svgscale}}%
    \fi%
  \else%
    \setlength{\unitlength}{\svgwidth}%
  \fi%
  \global\let\svgwidth\undefined%
  \global\let\svgscale\undefined%
  \makeatother%
  \begin{picture}(1,1.12418301)%
    \lineheight{1}%
    \setlength\tabcolsep{0pt}%
    \put(0,0){\includegraphics[width=\unitlength,page=1]{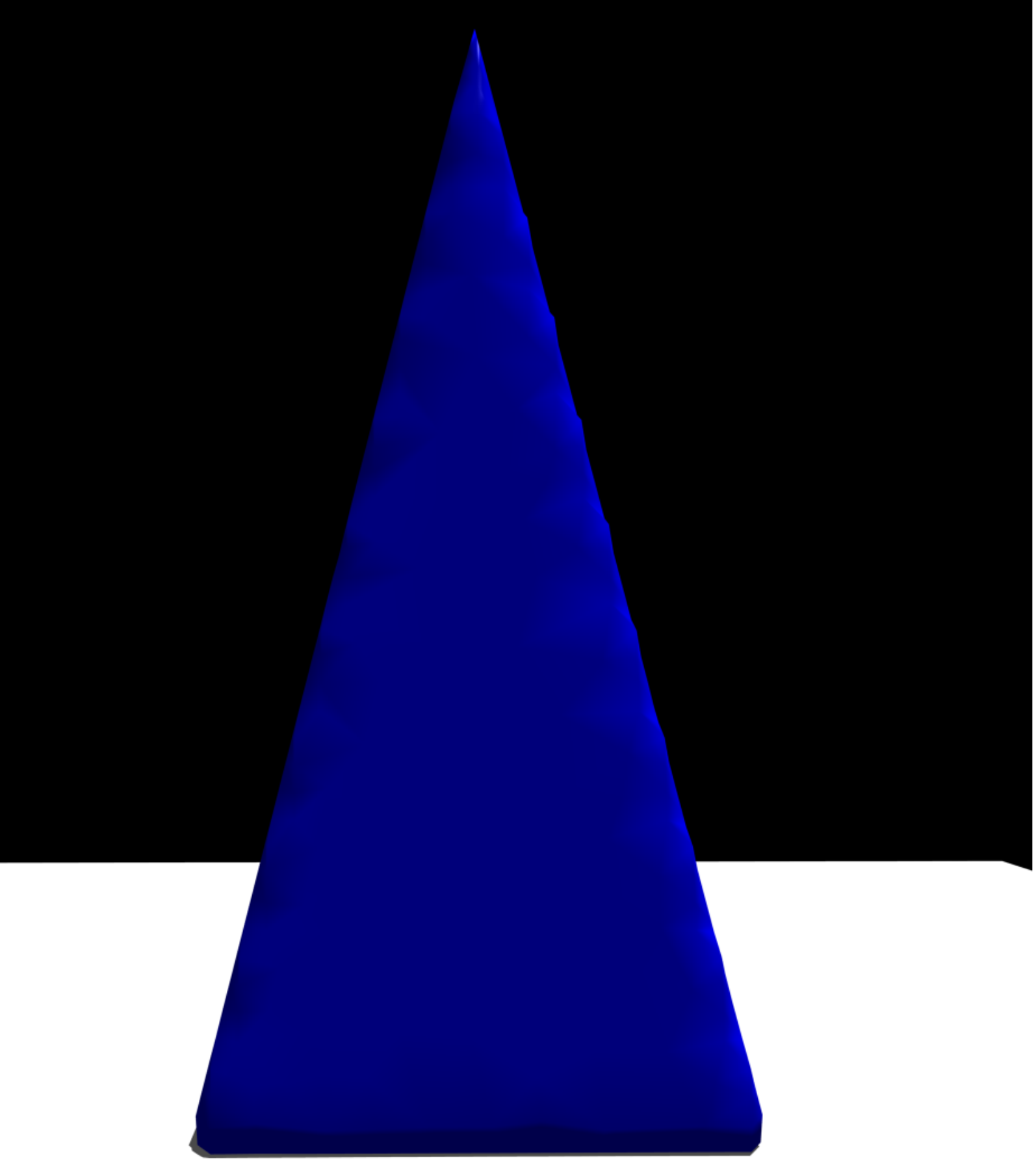}}%
    \put(0.7956094,0.91753466){\makebox(0,0)[t]{\lineheight{1.25}\smash{\begin{tabular}[t]{c}Grasp center\end{tabular}}}}%
    \put(0,0){\includegraphics[width=\unitlength,page=2]{grasprep.pdf}}%
    \put(0.80555007,1.02380852){\makebox(0,0)[t]{\lineheight{1.25}\smash{\begin{tabular}[t]{c}Grasp width\end{tabular}}}}%
    \put(0.79438273,0.81257537){\makebox(0,0)[t]{\lineheight{1.25}\smash{\begin{tabular}[t]{c}Grasp angle\end{tabular}}}}%
    \put(0.88367926,0.6389396){\makebox(0,0)[t]{\lineheight{1.25}\smash{\begin{tabular}[t]{c}Finger \\height\end{tabular}}}}%
  \end{picture}%
\endgroup%
}    
    \caption{Grasps are represented as rectangles defined by the center, angle, and width of the grasps}
    \label{fig:grasprep}
\end{figure}
\begin{figure}[ht]
    \centering
    \includegraphics[width=.9\linewidth]{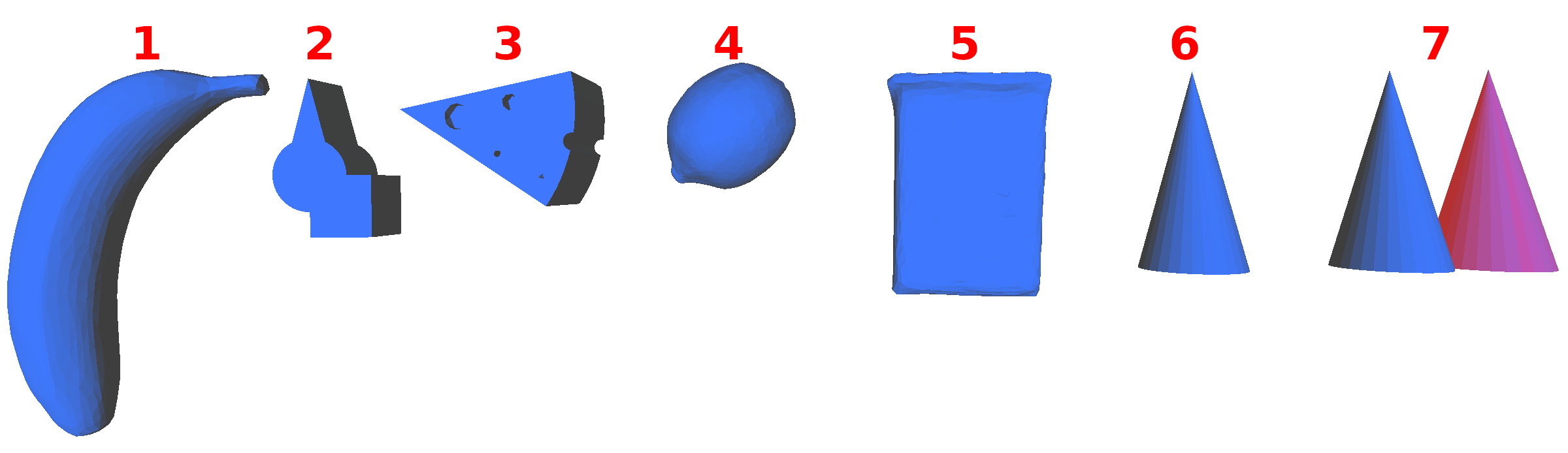}
    \caption{Seven common numbered objects used in the
experiment. All objects are single-material objects except for object 7, where the stiffness of its red part can be varied.}
    \label{fig:testobj}
\end{figure}

\begin{figure}[!t]
\centering
\subfloat[Generated grasps in the case of soft sponge (E = 2e4)]{
	\label{subfig:softsponge}
	\def\svgwidth{\linewidth}
         {\fontsize{8}{8}
        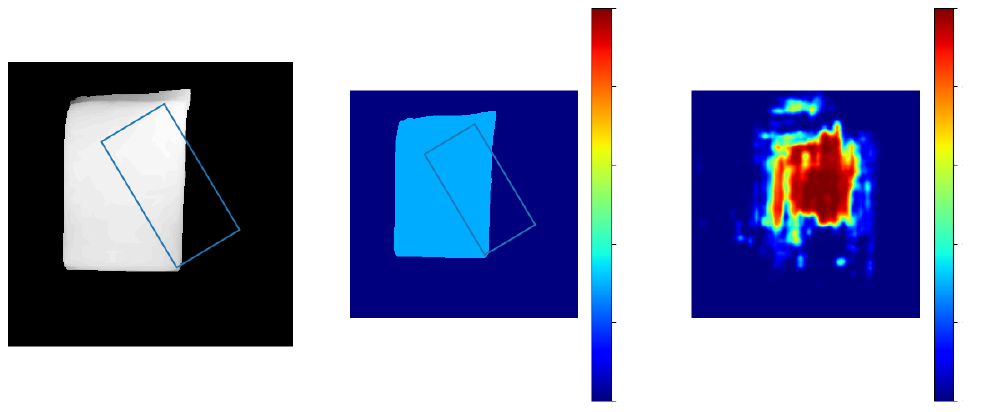}} \\
\subfloat[Generated grasps in the case of hard sponge (E = 2e9)]{
	\label{subfig:hardsponge}
		\def\svgwidth{\linewidth}
         {\fontsize{8}{8}
        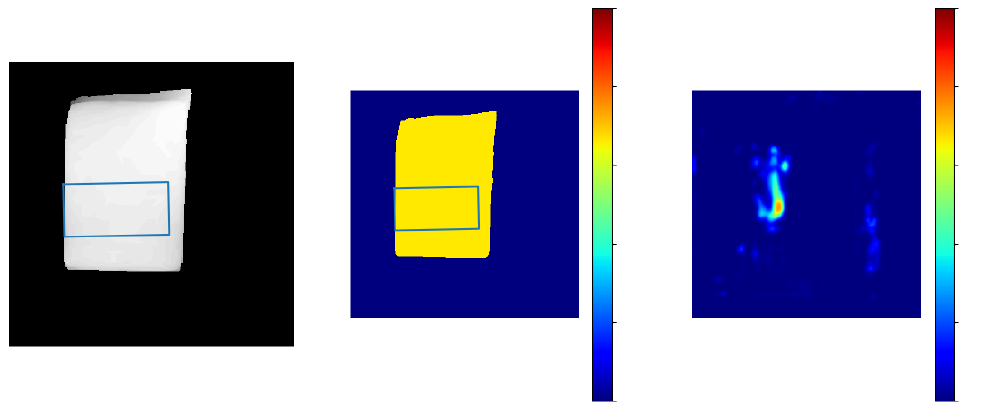} } 
\caption{In the case of soft sponge (a), the proposed method learned that the grasp quality is high across the whole objects thanks to their deformation, which in turn, enables pinch grasp. While in the case of hard sponge (b), the high quality grasp tends to be generated at the center of the object, and the grasp width is almost as big as the object in order to successfully cage the object.}
\label{fig:demo}
\end{figure}


\end{document}